\documentclass[10pt,twocolumn,letterpaper]{article}

\usepackage{iccv}
\usepackage{times}
\usepackage{epsfig}
\usepackage{graphicx}
\usepackage{amsmath}
\usepackage{amssymb}
\usepackage{pifont}

\usepackage{color}
\usepackage[dvipsnames,svgnames]{xcolor}
\usepackage{colortbl}
\definecolor{darkergreen}{RGB}{21, 152, 56}
\definecolor{red2}{RGB}{252, 54, 65}
\newcommand{\cmark}{\textcolor{darkergreen}{\ding{51}}}%
\usepackage{makecell}
\usepackage{tabularx}
\usepackage{multirow}
\usepackage{multicol}

\usepackage{siunitx}
\usepackage{cite}
\usepackage{ctable}
\usepackage{hhline}
\usepackage{lipsum}
\usepackage{booktabs}
\usepackage{csquotes}
\usepackage[accsupp]{axessibility}

\newcolumntype{L}[1]{>{\raggedright\let\newline\\\arraybackslash\hspace{0pt}}m{#1}}
\newcolumntype{C}[1]{>{\centering\let\newline\\\arraybackslash\hspace{0pt}}m{#1}}
\newcolumntype{R}[1]{>{\raggedleft\let\newline\\\arraybackslash\hspace{0pt}}m{#1}}

\usepackage[pagebackref=true,breaklinks=true,letterpaper=true,colorlinks,bookmarks=false]{hyperref}

\iccvfinalcopy 

\def\iccvPaperID{7220} 

\ificcvfinal\fi

\begin{document}

\title{Human Part-wise 3D Motion Context Learning \\ for Sign Language Recognition}

\author{
	Taeryung Lee$^{1}$ \qquad Yeonguk Oh$^{2}$ \qquad Kyoung Mu Lee$^{1,2}$ \\
	$^{1}$IPAI,  $^{2}$Dept. of ECE \& ASRI, Seoul National University, Seoul, Korea\\
	{\tt\small \{trlee94, namepllet, kyoungmu\}@snu.ac.kr} 
}


\maketitle
\ificcvfinal\thispagestyle{empty}\fi

\begin{abstract}
In this paper, we propose P3D, the human part-wise motion context learning framework for sign language recognition.
Our main contributions lie in two dimensions: learning the part-wise motion context and employing the pose ensemble to utilize 2D and 3D pose jointly.
First, our empirical observation implies that part-wise context encoding benefits the performance of sign language recognition.
While previous methods of sign language recognition learned motion context from the sequence of the entire pose, we argue that such methods cannot exploit part-specific motion context.
In order to utilize part-wise motion context, we propose the alternating combination of a part-wise encoding Transformer (PET) and a whole-body encoding Transformer (WET).
PET encodes the motion contexts from a part sequence, while WET merges them into a unified context.
By learning part-wise motion context, our P3D achieves superior performance on WLASL compared to previous state-of-the-art methods.
Second, our framework is the first to ensemble 2D and 3D poses for sign language recognition.
Since the 3D pose holds rich motion context and depth information to distinguish the words, our P3D outperformed the previous state-of-the-art methods employing a pose ensemble.
\end{abstract}


\section{Introduction}
Understanding and translating sign language is essential due to the huge potential benefit to society.
Sign language recognition (SLR) can be considered as a specific kind of action recognition (AR) since both tasks aim to classify human motion.
However, SLR has a characteristic that is distinguished from AR; Actions mainly focus on the motion of the human body, while the SLR system should be aware of more detailed information such as hand gestures and facial expressions.
Thus, a specified system for SLR will provide further gain over using the AR models on SLR.

\begin{figure}[t]
\begin{center}
\vspace{2mm}
\includegraphics[width=1.\linewidth]{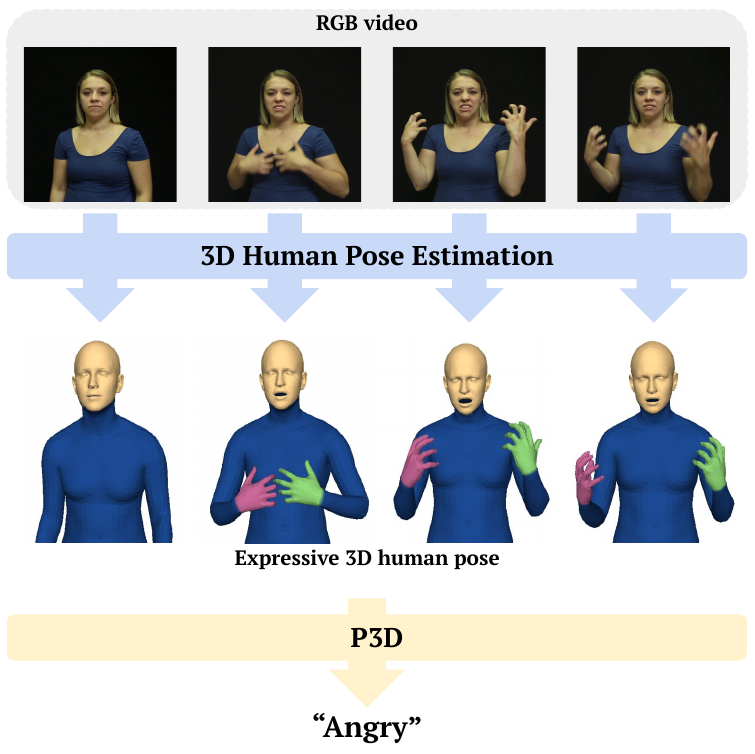}
\end{center}
\caption{\textbf{Overall pipeline of proposed system.}
We first extract expressive 3D human pose from RGB video using off-the-shelf pose estimation methods~\cite{cao2017realtime,moon2022accurate}.
The expressive 3D human pose consists of 2D pose, 3D pose and facial expression. 
Then our proposed P3D predicts the word from the expressive 3D human pose.}
\label{Overall_Pipeline}
\end{figure}

\begin{figure*}[t]
\begin{center}
\includegraphics[width=\linewidth]{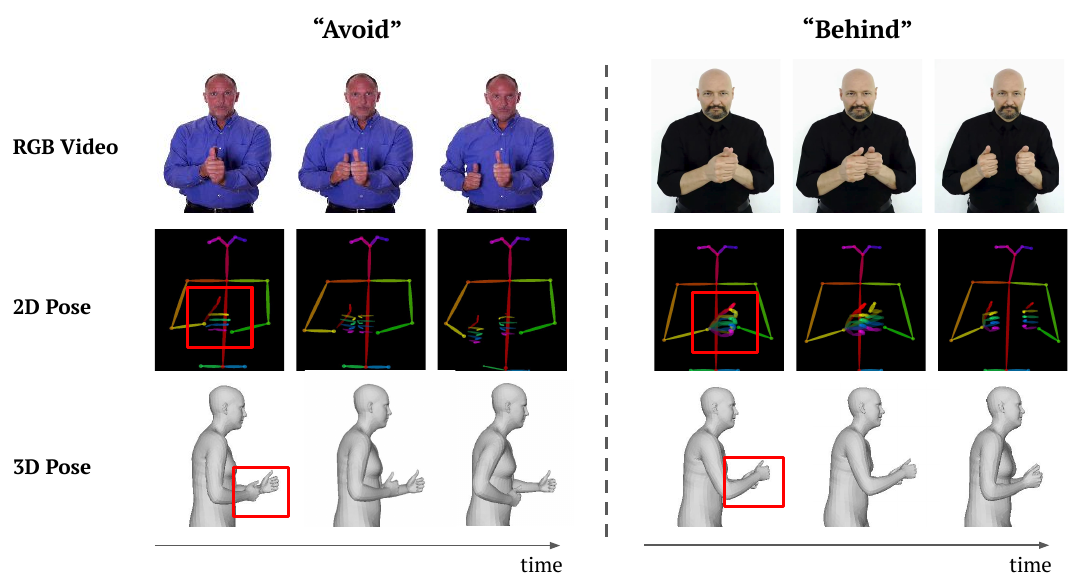}
\end{center}
\caption{\textbf{Effectiveness of 3D pose for sign language recognition.} 
The 2D poses of the words "Avoid" and "Behind" are almost identical.
Due to the absence of depth information, solely relying on 2D pose makes it hard to distinguish the words.
On the other hand, as shown in the red boxes, additionally employing the 3D pose with rich depth information can effectively resolve the ambiguity in sign language recognition.
}
\label{fig:3Deffectiveness}
\end{figure*}

Previous approaches on SLR problem are mainly categorized into RGB-based~\cite{camgoz2020sign, li2020transferring, albanie2020bsl} and pose-based methods~\cite{tunga2021pose, yan2018stgcn, song2017end, li2018co, koller2020survey}.
Pose-based methods are known to have the advantage over RGB-based approaches in terms of the robustness on domain gap~\cite{moon2021integralaction, duan2022revisiting}.
For that, we aimed to enhance the performance of SLR, while preserving the benefits inherent in the pose-based approach.

Although recent pose-based SLR methods have shown noticeable progress, there are two main limitations of existing methods.
First, they utilize only whole-body representation (\eg, skeleton) to understand the motion context of human~\cite{li2020word,tunga2021pose,hu2021signbert}.
Since such approaches overlook the detailed motion context within a specific part of human~\cite{guo2019human,li2022skeleton}, it leads to a decrease in SLR accuracy.
Second, existing pose-based methods only employ 2D pose~\cite{tunga2021pose, bohavcek2022spoter, li2020word, hu2021signbert}, which induces ambiguities in distinguishing the different words.
As shown in Figure~\ref{fig:3Deffectiveness}, the 2D pose of different words can be almost identical due to the absence of depth information.


To overcome the above limitations, we propose P3D, the human part-wise motion context learning framework for accurate sign language recognition.
The proposed P3D employs the part-wise encoding Transformer (PET) and whole-body encoding Transformer (WET) as main components.
Given the pose sequence, we first divide it into four parts: body, right hand, left hand, and face.
In contrast to previous methods where entire human poses are jointly processed, we utilize part-specific PET to encode the motion context of each part.
As each PET takes a pose sequence of a single part, it can learn more detailed motion context, which is pivotal for sign language recognition.
Then, the encoded features from each part are merged by WET.
While PET only considers the intra-part motion context, WET covers the entire part.
For the effective fusion of the inter-part and intra-part motion context, we alternately combined PET and WET inspired by spatio-temporal Transformers~\cite{arnab2021vivit, zhang2022mixste}.
As a result, our proposed P3D has outperformed the previous state-of-the-art~\cite{yan2018stgcn, bohavcek2022spoter} on WLASL.
Our ablation study on layer design supports part-wise motion context learning benefits SLR performance.
Especially, we observed that solely forming the model with WET (\ie, without part-wise motion context) degrades the performance, thus directly induces that our PET contributed to improving the accuracy.


We alleviate the second limitation of previous methods through the ensemble of 2D and 3D poses.
In terms of 3D pose, we utilize both 3D positional pose and 3D rotational pose (\eg, SMPL-X~\cite{SMPL-X:2019} pose parameters).
The 3D positional pose complements depth information that the 2D pose does not contain, and the 3D rotational pose provides rich motion context on the kinematic chain.
Our experiment shows that our P3D outperforms the previous methods by a large margin with using identical inputs; our model shows the superior result from only 2D pose, and also from ensemble of 2D and 3D pose.
Moreover, our framework is the first pose-based method that recorded a comparable score or even outperformed the RGB-based methods.
We claim our main contributions as follows:
\begin{itemize}

    \item We proposed P3D, the human part-wise motion context learning framework for sign language recognition.
    Our experiment supports that part-wise learning highly benefits recognition performance.
    As a result, our proposed P3D outperformed the previous SOTA on the WLASL benchmark.


    \item We employ the pose ensemble by joint-wise pose concatenation in order to exploit both 2D and 3D poses.
    We observed significant performance gain via pose ensemble, and P3D outperformed the previous methods by a large margin by jointly using 2D and 3D poses.

\end{itemize}

\section{Related works}
\subsection{Sign language recognition}
Word-level sign language recognition task aims to classify short-term videos by the corresponding word label.
Previous works on sign language recognition can be mainly categorized into two groups by the input modality: video-based and pose-based approaches.

\noindent\textbf{Video-based sign language recognition.} 
With the significant performance of the CNN-based action recognition, the majority of the prior works have designed their models based on CNN backbone~\cite{cheng2020fully, koller2018deep, zhou2021improving}.
Primary works~\cite{cui2017recurrent, cui2019deep, koller2019weakly, saunders2021continuous} first used convolutional neural networks to extract the frame-wise features.
Then, such methods encoded the temporal semantics along the time dimension using recurrent neural networks.
On the other hand, similar to action recognition, 3D CNNs are widely used in sign language recognition.
C3D~\cite{jia2014caffe, tran2015c3d, karpathy2014large} is the first 3D CNN model for action recognition, and it is widely used as a baseline model for quantitative comparison.
I3D~\cite{carreira2017i3d} is proposed based on the C3D~\cite{tran2015c3d} architecture and applied on sign language recognition in~\cite{joze2018ms}.
Li \textit{et al.}~\cite{li2020word} have compared various architectures, including 3D CNN and 2D CNN combined with RNN.
Hosain \textit{et al.}~\cite{hosain2021hand} proposed Fusion-3 model based on I3D Inception-v1~\cite{carreira2017i3d} with pose-guided pooling.

\noindent\textbf{Pose-based sign language recognition.}
Compared to the RGB video, the human pose sequence holds a compact and semantic-aware representation of human motion.
ST-GCN~\cite{yan2018stgcn} is the first pose-based approach to action recognition, utilizing the spatio-temporal graph convolutional network to encode the motion on kinetic chain.
Following works also have used the spatio-temporal architecture based on graph convolutional network~\cite{tunga2021pose, li2020word} and Transformers~\cite{vaswani2017attention}~\cite{bohavcek2022spoter, hu2021signbert}.

\subsection{3D human pose and shape estimation}
3D human pose and shape estimation aims to localize 3D coordinates of vertices of the human mesh.
Considering the various applications such as VR/AR and human action recognition, research on 3D human pose and shape estimation has been widely conducted.
Most works tried to reconstruct 3D human pose and shape for specific part of human, \eg body~\cite{moon2020i2l, choi2020pose2mesh, lin2021end}, hand~\cite{kulon2020weakly, chen2021i2uv}, or face~\cite{sanyal2019learning}.
Several works~\cite{pavlakos2019expressive, moon2022accurate} proposed methods that simultaneously estimate the 3D pose and shape for all human parts, \ie, whole-body.
By reconstructing 3D poses and shapes for all human parts, we can understand human intention and feelings more precisely.
In this work, we leverage Hand4Whole~\cite{moon2022accurate} to acquire a 3D pose of all human parts from RGB videos.
To the best of our knowledge, we first utilize 3D pose for sign language recognition. 

\subsection{Transformers for human motion.}
After the remarkable success of Transformers in natural language processing (NLP)~\cite{vaswani2017attention}, vision researchers have employed Transformers for various vision tasks~\cite{carion2020end,dosovitskiy2020image,lin2021end}.
Transformers also have been widely adopted for tasks that require human motion understanding, such as 2D-to-3D pose sequence lifting~\cite{zheng20213d,zhang2022mixste,li2022mhformer} and skeleton-based action recognition~\cite{plizzari2021spatial,gao2022focal}.
In the field of 2D-to-3D pose sequence lifting, PoseFormer~\cite{zheng20213d} proposed a pure Transformer-based approach for 2D-to-3D pose sequence lifting.
MixSTE~\cite{zhang2022mixste} utilized separated joint-wise temporal Transformer and spatial Transformer alternately.
MHFormer~\cite{li2022mhformer} employed Transformers to generate and refine multiple hypotheses for 3D pose sequences from 2D pose sequences.
In the field of skeleton-based action recognition, ST-TR~\cite{plizzari2021spatial} proposed Transformer-based two-stream networks where each stream learns spatial and temporal correlations in a skeleton sequence.
FG-STFormer~\cite{gao2022focal} presented a variation of vanilla Transformer to capture local and global information effectively. 
Our P3D also leverages Transformer-based modules to learn the human motion.
However, most previous works only focused on the motion of the human body, while proposed P3D learns the motion of all human parts, including hands and facial expressions.

\begin{figure*}[t]
\begin{center}
\includegraphics[width=\linewidth]{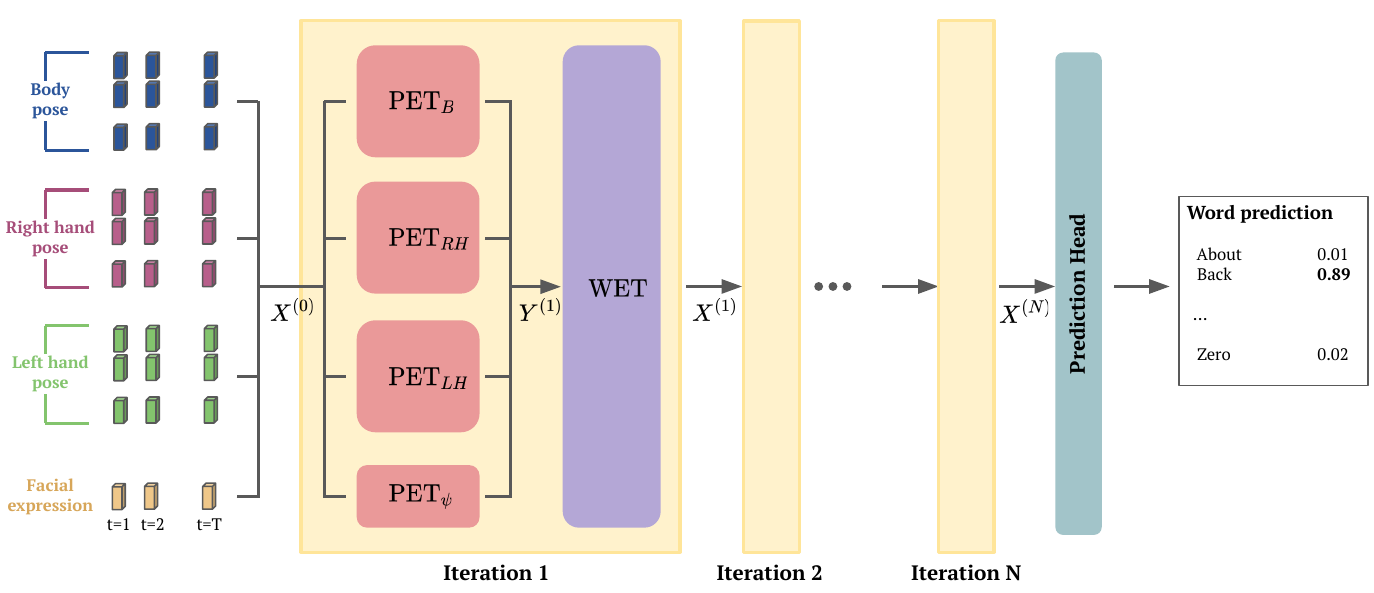}
\end{center}
\caption{\textbf{Overall architecture of P3D.} 
Our P3D aims to predict the word label corresponding to a given pose sequence.
We first provide joint-wise linearly embedded feature vectors as an input, along with linearly encoded facial expression features.
P3D consists of 2$N$ alternating layers of PET, which are assigned for each part, and WET, which covers entire parts.
After the encoder, the prediction head takes the mean along the time dimension, then sequentially passes through batch normalization, linear layer, and softmax layer to produce the word prediction.
}
\label{fig:Overall_Architecture}
\end{figure*}

\begin{figure}[t]
\begin{center}
\includegraphics[width=1.\linewidth]{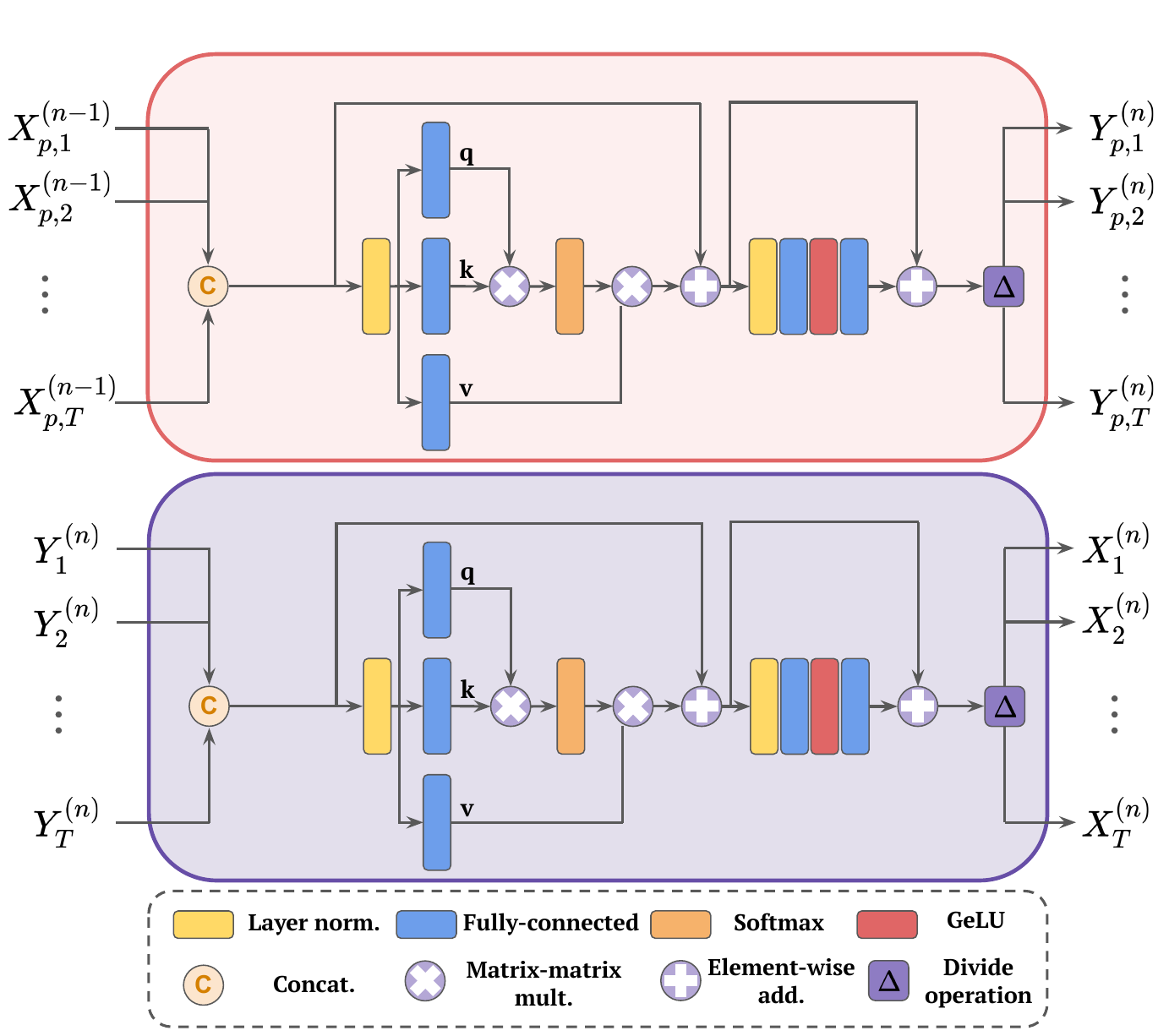}
\end{center}
\caption{\textbf{Overall architectures of PET and WET.}
We adapt the Transformer encoder layer~\cite{vaswani2017attention} with vanilla self-attention for both PET (above) and WET (below).
For the PET, the subscript $p \in \{\text{B}, \text{RH}, \text{LH}, \psi\}$ denotes the assigned human part. 
}
\label{fig:PET_WET}
\end{figure}

\section{Method}

\subsection{Overview}
In this section, we describe the input data format and architecture of our P3D model.
Figure~\ref{fig:Overall_Architecture} illustrates the overview of our model.
We provide a sequence of human poses to our model as an input
P3D leverages alternating structure of Transformer layers~\cite{vaswani2017attention}, inspired by the spatio-temporal architectures from the previous works~\cite{arnab2021vivit, zhang2022mixste}.
The first layer, the part-wise encoding Transformer (PET), is designed to learn intra-part motion context.
To capture the specific motion context from each part, we assign the dedicated Transformers.
Meanwhile, the second whole-body encoding Transformer (WET) encodes the inter-part motion context, taking account of every human part used in the model.
As the last step, we take mean along the time axis on encoded representations, then sequentially pass through batch normalization, linear layer, and softmax layer to get the final output.

\begin{figure*}[t]
\begin{center}
\includegraphics[width=\linewidth]{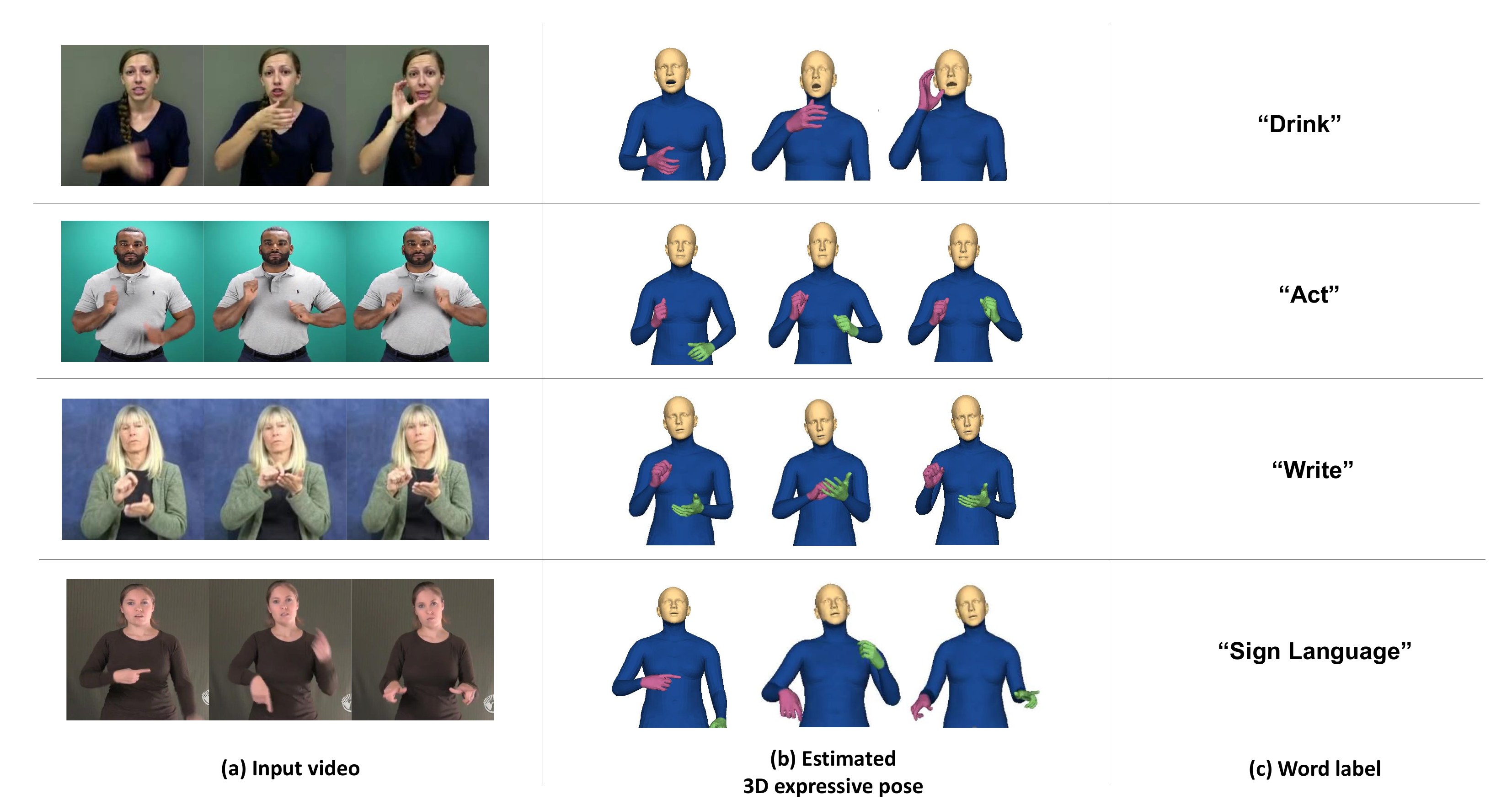}
\end{center}
\caption{\textbf{Visualization.} 
The 3D expressive pose of the signer (b) is estimated from the given sign language video (a) with off-the-shelf estimator Hand4Whole~\cite{moon2022accurate}.
The corresponding word label (c) is denoted in the right. 
}
\label{fig:qualitative_result}
\end{figure*}

\subsection{Input format}

\noindent\textbf{Human motion representation.}
The overview of the input data structure is illustrated on the left of Figure~\ref{fig:Overall_Architecture}.
As a pose-based sign language model, our P3D takes a sequence $(X_1, ..., X_T)\in\mathbb{R}^{T \times (J \times F + E)}$ of $T$ human poses as an input.
Pose representation $X_t = (\mathbf{x}_1, ..., \mathbf{x}_J, e) \in \mathbb{R}^{J \times F + E}$ for each time $t \in \{1, 2, ..., T\}$ consists of joint-wise pose representation $\mathbf{x}_j \in \mathbb{R}^F$ for each joint $j \in \{1, 2, ..., J\}$, and facial expression feature $e \in \mathbb{R}^E$~\cite{SMPL-X:2019}.
From our experiment on the pose fusion method, we have observed that joint-wise concatenation of the 2D positional pose (2D pos.), 3D positional pose (3D pos.), and 3D rotational pose (3D rot.) performed the best on WLASL2000.
Thus, we form the joint-wise pose representation $\mathbf{x}_j = (\mathbf{x}_{j, \text{2D pos}}, \mathbf{x}_{j, \text{3D pos}}, \mathbf{x}_{j, \text{3D rot}}) \in \mathbb{R}^F$ where $\mathbf{x}_{j, \text{2D pos}} \in \mathbb{R}^2$, $\mathbf{x}_{j, \text{3D pos}} \in \mathbb{R}^3$, $\mathbf{x}_{j, \text{3D rot}} \in \mathbb{R}^6$ represents the 2D positional pose, 3D positional pose and 3D rotational pose~\cite{zhou2019rotation6d} of joint $j$, respectively.
Please refer to our experiment on Section 4.3 for the details of pose fusion methods.
Following the literature~\cite{hu2021signbert}, we use $T=32$ for both training and testing.

\noindent\textbf{Joint set and human part.}
We use $J=40$ joints and 4 human parts to represent the human motion of the sign language.
Ten joints above the pelvis are selected from the SMPL~\cite{SMPL:2015} joint set, defined as a \textit{body} part.
Fifteen joints, three per finger, represent each \textit{hands}.
For the \textit{face}, we use the 10-dimensional feature vector from SMPL-X~\cite{SMPL-X:2019} to represent the \textit{facial expression}.
Please refer to Section 4.2 for the details of human pose preprocessing.

\subsection{Architecture}
\noindent\textbf{Linear layer.}
Our model first linearly encodes each joint-wise pose representation $\mathbf{x}_j \in \mathbb{R}^{F}$ into D dimensional vector.
We also assign a dedicated linear layer to encode facial expressions into $\alpha D$ dimensional vector.
Empirically, we decided to use $D = 8$ and $\alpha = 10$.
As a result, pose $X_t = (\mathbf{x}_1, ..., \mathbf{x}_J, e) \in \mathbb{R}^{J \times F + E}$ at time $t$ is encoded into $X^{(0)}_t \in \mathbb{R}^{(J \times D + \alpha D) = ((10 + 15 + 15) \times 8 + 80)}$.
Note that 10, 15, and 15 represent the number of joints in the body, left hand and right hand, and thus can be separated into each human part via indexing.
Finally we form the input $X^{(0)} = (X^{(0)}_1, X^{(0)}_2, ..., X^{(0)}_T) \in \mathbb{R}^{T \times ((10 + 15 + 15) \times 8 + 80)}$ with positional encoding~\cite{vaswani2017attention}.

\noindent\textbf{Alternating Transformers.}
We adapted the idea of alternating two types of layers~\cite{arnab2021vivit, zhang2022mixste}, which have different purposes and expertise.
Our P3D consists of $2N$ alternating layers: $N$ Part-wise Encoding Transformer (PET) and $N$ Whole-body Encoding Transformer (WET).
Starting from linear encoded input $X^{(0)}$, we denote the output of $i$'th PET as $Y^{(i)}$ and the output of $i$'th WET as $X^{(i)}$.
As the dimension remains identical through Transformers, our features $X^{(i)}$ and $Y^{(i)}$ remains in a same dimension as $X^{(0)}$.
Empirically, we use $N=3$ layers for WLASL.

\noindent\textbf{Part-wise encoding Transformer (PET).}
The architecture of PET is illustrated in the upper part of Figure~\ref{fig:PET_WET}.
From previous layer, PET takes 2D tensor $X^{(i)} \in \mathbb{R}^{T \times ((10 + 15 + 15) \times 8 + 80)}$ as an input.
The first (horizontal) and second (vertical) dimensions represent the time and feature axis, respectively.
We first separate features for each human part $X^{(i)}_{B} \in \mathbb{R}^{T \times 80}$, $X^{(i)}_{LH} \in \mathbb{R}^{T \times 120}$, $X^{(i)}_{RH} \in \mathbb{R}^{T \times 120}$, $X^{(i)}_\psi \in \mathbb{R}^{T \times 80}$ via slicing on second (feature) dimension by indices (80, 200, 320, 400), respectively.
Then we pass each feature through dedicated Transformer encoder layers PET$_{B}$, PET$_{RH}$, PET$_{LH}$, and PET$_\psi$ assigned for each human part, respectively.
Finally, we concatenate the outputs along the second (feature) dimension into a single 2D tensor $Y^{(i)} \in \mathbb{R}^{T \times ((10 + 15 + 15) \times 8 + 80)}$.

\noindent\textbf{Whole-body encoding Transformer (WET).}
The architecture of WET is illustrated in the lower part of Figure~\ref{fig:PET_WET}.
From the previous layer PET, WET takes 2D tensor $Y^{(i)} \in \mathbb{R}^{T \times ((10 + 15 + 15) \times 8 + 80)}$ as an input.
Since WET aims to learn inter-part motion context, we assign a single Transformer to encode the whole motion sequence.
Output of WET $X^{(i+1)} \in \mathbb{R}^{T \times ((10 + 15 + 15) \times 8 + 80)}$ is passed to next layer without modification.

\noindent\textbf{Prediction head.}
From Transformer-encoded feature $X^{(N)} \in \mathbb{R}^{T \times 400}$, we first calculate the mean feature vector $\mathbf{x}_\text{feat} \in \mathbb{R}^{400}$ by taking the mean along the first (time) dimension.
Then, we pass $\mathbf{x}_\text{feat}$ sequentially through 1D batch normalization, a single FC-layer, and a softmax layer.
As a result, our model produces the final prediction $\mathbf{x}_\text{class} \in \mathbb{R}^C$, where $C$ is the number of target classes (glosses) in the dataset.

\noindent\textbf{Training and inference.}
We train P3D with a cross-entropy loss between predicted $\mathbf{x}_\text{class}$ and one-hot encoded GT class vector.
We randomly sample continuous $T$ frames to form the training inputs.
During the inference phase, we sample four chunks of $T$ continuous frames from the human pose sequence in the test set following the previous work~\cite{li2020word}.
Then, we use P3D to generate the class prediction $\mathbf{x}_\text{class}$ for each $T$ continuous frame, and average them out to produce the final prediction for the entire sequence.

\begin{table*}
\footnotesize
\centering
\setlength\tabcolsep{1.0pt}
\def\arraystretch{1.1}
\begin{tabular}
{L{3.0cm}|C{1.05cm}C{1.05cm}|C{1.05cm}C{1.05cm}|C{1.05cm}C{1.05cm}|C{1.05cm}C{1.05cm}|C{1.05cm}C{1.05cm}|C{1.05cm}C{1.05cm}}
\specialrule{.1em}{.05em}{.05em}
\hline
\multirow{3}{*}{Pose encoding methods} & \multicolumn{4}{c|}{WLASL 100} & \multicolumn{4}{c|}{WLASL 300} & \multicolumn{4}{c}{WLASL 2000} \\
\cline{2-13}
& \multicolumn{2}{c|}{Per-instance} & \multicolumn{2}{c|}{Per-class} & \multicolumn{2}{c|}{Per-instance} & \multicolumn{2}{c|}{Per-class} & \multicolumn{2}{c|}{Per-instance} & \multicolumn{2}{c}{Per-class} \\
& Top-1 & Top-5 & Top-1 & Top-5 & Top-1 & Top-5 & Top-1 & Top-5 & Top-1 & Top-5 & Top-1 & Top-5 \\
\hline
MLP-WET & 71.89 & 85.94 & 72.33 & 87.21 & 59.44 & 82.20 & 59.70 & 82.43 & 34.97 & 70.57 & 32.78 & 69.41 \\
MLP-PET & 72.28 & 88.76 & 73.76 & 89.55 & 57.89 & 82.50 & 58.56 & 82.96 & 34.72 & 70.71 & 32.49 & 69.66 \\
PET-PET & \textbf{77.91} & \textbf{94.77} & \textbf{78.30} & \textbf{95.25} & \textbf{67.33} & 88.68 & \textbf{68.06} & 88.06 & 32.39 & 63.61 & 30.34 & 62.20 \\
WET-WET & 75.50 & 91.57 & 76.43 & 92.68 & 66.87 & 88.24 & 66.92 & 88.56 & 40.72 & 74.70 & 38.26 & 73.54 \\
\textbf{PET-WET (Ours)} & 76.71 & 91.97 & 78.27 & 92.97 & 67.18 & \textbf{89.01} & 67.62 & \textbf{89.24} & \textbf{44.47} & \textbf{79.97} & \textbf{42.18} & \textbf{78.52} \\
\hline
\specialrule{.1em}{.05em}{.05em}
\end{tabular}
\vspace*{3mm}
\caption{\textbf{Ablation study on pose encoding methods.}
We ablate the performance of pose encoding methods in single iteration of P3D.
Our part-wise context learning (PET-WET) scored the best performance.
The observation that ours (PET-WET) has outperformed WET-WET (Transformer without part) directly supports our main contribution of part-wise context learning.
}
\label{table:pose_encoding_methods}
\end{table*}

\begin{table*}
\footnotesize
\centering
\setlength\tabcolsep{1.0pt}
\def\arraystretch{1.1}
\begin{tabular}
{C{1.0cm}C{1.0cm}C{1.0cm}|C{1.05cm}C{1.05cm}|C{1.05cm}C{1.05cm}|C{1.05cm}C{1.05cm}|C{1.05cm}C{1.05cm}|C{1.05cm}C{1.05cm}|C{1.05cm}C{1.05cm}}
\specialrule{.1em}{.05em}{.05em}
\hline
\multicolumn{3}{c|}{\multirow{2}{*}{Input human parts}} & \multicolumn{4}{c|}{WLASL 100} & \multicolumn{4}{c|}{WLASL 300} & \multicolumn{4}{c}{WLASL 2000} \\
\cline{4-15}
& & & \multicolumn{2}{c|}{Per-instance} & \multicolumn{2}{c|}{Per-class} & \multicolumn{2}{c|}{Per-instance} & \multicolumn{2}{c|}{Per-class} & \multicolumn{2}{c|}{Per-instance} & \multicolumn{2}{c}{Per-class} \\
 Body & Hands & Expr. & Top-1 & Top-5 & Top-1 & Top-5 & Top-1 & Top-5 & Top-1 & Top-5 & Top-1 & Top-5 & Top-1 & Top-5 \\
\hline
\cmark & & & 44.98 & 74.70 & 44.75 & 75.85 & 37.62 & 70.43 & 37.40 & 70.17 & 17.80 & 40.76 & 15.70 & 37.57 \\
\cmark & & \cmark& 46.99 & 72.29 & 45.57 & 72.43 & 38.70 & 70.43 & 38.73 & 70.45 & 20.98 & 47.05 & 18.73 & 43.94 \\
 & \cmark & & 71.89 & 91.16 & 73.10 & 92.08 & 62.07 & 87.15 & 62.69 & 87.93 & 39.00 & 73.40 & 36.97 & 72.16 \\
\cmark & \cmark & & 75.90 & 91.56 & 78.10 & 91.97 & 66.25 & \textbf{89.16} & 65.92 & \textbf{89.52} & 42.35 & 77.15 & 40.74 & 76.67 \\
\cmark & \cmark & \cmark & \textbf{76.71} & \textbf{91.97} & \textbf{78.27} & \textbf{92.97} & \textbf{67.18} & 89.01 & \textbf{67.62} & 89.24 & \textbf{44.47} & \textbf{79.97} & \textbf{42.18} & \textbf{78.52} \\
\hline
\specialrule{.1em}{.05em}{.05em}
\end{tabular}
\vspace*{3mm}
\caption{\textbf{Ablation study on input human parts for P3D.}
We ablate the performance of provided input human parts. Check marks in the first column implies the given input part. We abbreviate the facial expression to Expr.
}
\label{table:input_human_parts}
\end{table*}


\begin{table*}
\footnotesize
\centering
\setlength\tabcolsep{1.0pt}
\def\arraystretch{1.1}
\begin{tabular}
{C{1.0cm}C{1.0cm}C{1.0cm}|C{1.05cm}C{1.05cm}|C{1.05cm}C{1.05cm}|C{1.05cm}C{1.05cm}|C{1.05cm}C{1.05cm}|C{1.05cm}C{1.05cm}|C{1.05cm}C{1.05cm}}
\specialrule{.1em}{.05em}{.05em}
\hline
\multicolumn{3}{c|}{\multirow{2}{*}{Input pose representations}} & \multicolumn{4}{c|}{WLASL 100} & \multicolumn{4}{c|}{WLASL 300} & \multicolumn{4}{c}{WLASL 2000} \\
\cline{4-15}
& & & \multicolumn{2}{c|}{Per-instance} & \multicolumn{2}{c|}{Per-class} & \multicolumn{2}{c|}{Per-instance} & \multicolumn{2}{c|}{Per-class} & \multicolumn{2}{c|}{Per-instance} & \multicolumn{2}{c}{Per-class} \\
2D pos. & 3D pos. & 3D rot. & Top-1 & Top-5 & Top-1 & Top-5 & Top-1 & Top-5 & Top-1 & Top-5 & Top-1 & Top-5 & Top-1 & Top-5 \\
\hline
\cmark & & & 60.24 & 83.13 & 60.45 & 82.63 & 52.17 & 79.72 & 52.59 & 79.79 & 35.99 & 67.26 & 33.47 & 65.15 \\
& \cmark & & 41.09 & 66.67 & 40.98 & 67.02 & 39.97 & 69.76 & 40.01 & 70.01 & 31.69 & 62.82 & 28.42 & 59.83 \\
& & \cmark & 72.87 & 88.37 & 73.60 & 88.93 & 61.68 & 82.34 & 61.93 & 82.78 & 38.60 & 73.42 & 35.57 & 71.86 \\
\cmark & \cmark & & 55.02 & 81.12 & 55.73 & 82.01 & 50.15 & 80.19 & 49.74 & 81.09 & 36.91 & 69.09 & 34.00 & 67.06 \\
\cmark & & \cmark & 75.50 & 91.16 & 76.88 & 91.58 & 65.94 & 88.85 & 66.26 & 88.75 & 43.48 & 77.64 & 41.39 & 76.38 \\
& \cmark & \cmark & 74.42 & 88.37 & 75.15 & 89.26 & 62.72 & 83.53 & 63.43 & 84.01 & 41.14 & 76.79 & 38.47 & 75.34 \\
\cmark & \cmark & \cmark & \textbf{76.71} & \textbf{91.97} & \textbf{78.27} & \textbf{92.97} & \textbf{67.18} & \textbf{89.01} & \textbf{67.62} & \textbf{89.24} & \textbf{44.47} & \textbf{79.69} & \textbf{42.18} & \textbf{78.52} \\
\hline
\specialrule{.1em}{.05em}{.05em}
\end{tabular}
\vspace*{3mm}
\caption{\textbf{Ablation study on input pose representations.}
We ablate the performance of provided input human pose representations. 
Check marks in the first column implies the given input pose representation.
Employing every pose representation scored the best result.
}
\label{table:pose_representations}
\end{table*}

\begin{table*}
\footnotesize
\centering
\setlength\tabcolsep{1.0pt}
\def\arraystretch{1.1}
\begin{tabular}
{C{3.0cm}|C{1.05cm}C{1.05cm}|C{1.05cm}C{1.05cm}|C{1.05cm}C{1.05cm}|C{1.05cm}C{1.05cm}|C{1.05cm}C{1.05cm}|C{1.05cm}C{1.05cm}}
\specialrule{.1em}{.05em}{.05em}
\hline
\multirow{3}{*}{Ensemble methods} & \multicolumn{4}{c|}{WLASL 100} & \multicolumn{4}{c|}{WLASL 300} & \multicolumn{4}{c}{WLASL 2000} \\
\cline{2-13}
& \multicolumn{2}{c|}{Per-instance} & \multicolumn{2}{c|}{Per-class} & \multicolumn{2}{c|}{Per-instance} & \multicolumn{2}{c|}{Per-class} & \multicolumn{2}{c|}{Per-instance} & \multicolumn{2}{c}{Per-class} \\
& Top-1 & Top-5 & Top-1 & Top-5 & Top-1 & Top-5 & Top-1 & Top-5 & Top-1 & Top-5 & Top-1 & Top-5 \\
\hline
Late & 77.90 & 93.17 & 78.30 & \textbf{95.25} & 67.34 & 88.23 & \textbf{68.06} & 88.68 & 32.71 & 64.04 & 30.34 & 63.67 \\
Middle & \textbf{79.92} & \textbf{93.57} & \textbf{81.66} & 94.38 & \textbf{67.80} & \textbf{89.94} & 67.86 & \textbf{90.13} & 42.63 & 71.71 & 40.17 & 76.43 \\
\textbf{Early (Ours)} & 76.71 & 91.97 & 78.27 & 92.97 & 67.18 & 89.01 & 67.62 & 89.24 & \textbf{44.47} & \textbf{79.69} & \textbf{42.18} & \textbf{78.52} \\
\hline
\specialrule{.1em}{.05em}{.05em}
\end{tabular}
\vspace*{3mm}
\caption{\textbf{Ablation study on pose representation ensemble methods.}
We report the quantitative result on pose representation ensemble methods.
Early ensemble by joint-wise concatenation outperforms the middle (between encoder \& head) and late (after head) ensemble.
}
\label{table:pose_fusion}
\end{table*}

\begin{table}
\begin{center}
\setlength\tabcolsep{1.0pt}
\def\arraystretch{1.1}
\begin{tabular}{C{2cm}|C{3cm}|C{3cm}}
\hline
\specialrule{.1em}{.05em}{.05em}
Ensemble methods & FLOPs (G) & \# Params (M) \\
\hline
Late & 182.06 & 10.71\\
Middle & 182.02 & 10.65\\
\textbf{Early (Ours)} & \textbf{83.66} & \textbf{4.94}\\
\hline
\specialrule{.1em}{.05em}{.05em}
\end{tabular}
\end{center}
\caption{\textbf{Computational costs of pose ensemble methods.} We used the FLOPs counter for PyTorch~\cite{paszke2017automatic} from fvcore package to measure the computational costs of the models.}
\label{table:pose_ensemble_flops}
\end{table}

\begin{table*}[t]
\footnotesize
\centering
\setlength\tabcolsep{1.0pt}
\def\arraystretch{1.1}
\begin{tabular}
{L{2.8cm}|C{1.1cm}C{1.1cm}|C{1.1cm}C{1.1cm}|C{1.1cm}C{1.1cm}|C{1.1cm}C{1.1cm}|C{1.1cm}C{1.1cm}|C{1.1cm}C{1.1cm}}
\specialrule{.1em}{.05em}{.05em}
\hline
\multirow{3}{*}{Methods} & \multicolumn{4}{c|}{WLASL 100} & \multicolumn{4}{c|}{WLASL 300} & \multicolumn{4}{c}{WLASL 2000} \\
\cline{2-13}
& \multicolumn{2}{c|}{Per-instance} & \multicolumn{2}{c|}{Per-class} & \multicolumn{2}{c|}{Per-instance} & \multicolumn{2}{c|}{Per-class} & \multicolumn{2}{c|}{Per-instance} & \multicolumn{2}{c}{Per-class} \\
& Top-1 & Top-5 & Top-1 & Top-5 & Top-1 & Top-5 & Top-1 & Top-5 & Top-1 & Top-5 & Top-1 & Top-5 \\
\hline
\textbf{\textit{RGB-based}} &   &   &   &   &   &   &   &   &   &   &   & \\
TK-3D ConvNet~\cite{ge20173d} & \textbf{77.55} & - & - & - & \textbf{68.75} & - & - & - & - & - & - & - \\
Fusion-3~\cite{hosain2021hand} & 75.67 & - & - & - & 68.3 & - & - & - & - & - & - & - \\
I3D~\cite{carreira2017i3d} & 65.89 & \textbf{84.11} & \textbf{67.01} & \textbf{84.58} & 56.14 & \textbf{79.94} & \textbf{56.24} & \textbf{78.38} & \textbf{32.48} & \textbf{57.31} & - & - \\
\hline
\hline
\textbf{\textit{Pose-based (2D)}} &   &   &   &   &   &   &   &   &   &   &   & \\
Pose-TGCN~\cite{li2020word} & 55.43 & 78.68 & - & - & 38.42 & 67.51 & - & - & 23.65 & 51.75 & - & - \\
GCN-BERT~\cite{duan2022posec3d} & 60.15 & 83.98 & - & - & 42.18 & 71.71 & - & - & - & - & - & - \\
SPOTER~\cite{bohavcek2022spoter} & \textbf{63.18} & - & - & - & 43.18 & - & - & - & 28.86 & 61.78 & 28.21 & 60.26 \\
ST-GCN~\cite{yan2018stgcn} & 50.78 & 79.07 & 51.62 & 79.47 & 44.46 & 73.05 & 45.29 & 73.16 & 34.40 & 66.47 & 32.53 & 65.45 \\
\textbf{P3D (Ours)} & 60.24 & \textbf{85.94} & \textbf{60.48} & \textbf{85.66} & \textbf{50.93} & \textbf{77.71} & \textbf{50.81} & \textbf{78.55} & \textbf{35.96} & \textbf{68.31} & \textbf{33.22} & \textbf{66.20} \\
\hline
\hline
\textbf{\textit{Pose-based (2D + 3D)}} &   &   &   &   &   &   &   &   &   &   &   & \\
SPOTER~\cite{bohavcek2022spoter} & 73.90 & 91.57 & 74.51 & 91.55 & 54.02 & 82.20 & 55.28 & 82.64 & 30.91 & 64.25 & 29.88 & 62.7 \\
ST-GCN~\cite{yan2018stgcn} & 56.63 & 79.51 & 56.93 & 79.71 & 45.82 & 76.01 & 45.96 & 76.22 & 37.37 & 72.13 & 35.61 & 70.66 \\
\textbf{P3D (Ours)} & \textbf{75.90} & \textbf{91.56} & \textbf{78.10} & \textbf{91.97} & \textbf{66.25} & \textbf{89.16} & \textbf{65.92} & \textbf{89.52} & \textbf{42.35} & \textbf{77.15} & \textbf{40.74} & \textbf{76.67} \\

\hline
\hline
\textbf{\textit{Pose-based (2D + 3D + facial expression)}} &   &   &   &   &   &   &   &   &   &   &   & \\
\textbf{P3D (Ours)} & \textbf{76.71} & \textbf{91.97} & \textbf{78.27} & \textbf{92.97} & \textbf{67.18} & \textbf{89.01} & \textbf{67.62} & \textbf{89.24} & \textbf{44.47} & \textbf{79.69} & \textbf{42.18} & \textbf{78.52} \\
\hline
\specialrule{.1em}{.05em}{.05em}
\end{tabular}
\vspace*{3mm}
\caption{\textbf{Comparisons with state-ot-the-art methods.}
We report the recognition accuracy of previous methods for comparison with ours.
The results in each block are trained and evaluated with using exact same input denoted in the first column.
Results in the first block show the recognition accuracy of existing RGB-based approaches.
The second block represents the pose-based approach with 2D pose inputs.
Our result in the second block is also generated solely with 2D poses.
Scores in the third block are obtained from employing pose ensemble on previous methods and ours, using both 2D and 3D poses.
Note that previous methods did not utilize 3D pose in their works, thus we applied pose ensemble based on released codes.
Finally, the fourth block shows the best performance of our model using 2D, 3D poses and facial expression.
Blank scores denoted with a hyphen are not reported in previous works.
}
\label{table:sota_comparision}
\end{table*}

\section{Experiment}
\subsection{Implementation details}
We used PyTorch~\cite{paszke2017automatic} Transformer implementation for our PET and WET, with 256 inner dimensions, 0.3 dropout rate, GELU activation layer, and 4 multi-head attentions.
We trained our model for 500 epochs using Adam~\cite{adam} optimizer with a mini-batch size of 512, a learning rate of 5$\times10^{-4}$, the weight decay rate of $5\times10^{-3}$.

\subsection{Datasets, evaluation metrics and preprocessing}
\noindent\textbf{WLASL dataset.} WLASL~\cite{li2020word} is a large-scale word-level sign language dataset that contains 21,083 videos with 119 signers.
The videos in WLASL are collected from the Internet, and each video depicts a signer performing a single sign.
The WLASL dataset consists of 2,000 words, and it also provides subsets of the dataset, WLASL100, and WLASL300.
Nevertheless, our main target is to push through the limit of recognition performance on WLASL2000.

\noindent\textbf{Evaluation metrics}
For the evaluation metrics, we employ Top-1 and Top-5 accuracy.
Following the previous work~\cite{hu2021signbert}, we calculate both metrics under two different settings, \ie, per-instance and per-class.
In per-instance setting, we measure the percentage of accurately classified video over the entire test instances.
Per-class setting is also utilized since each class (gloss) has a different number of instances.
In a per-class setting, we calculate the accuracy of each class and average them over all classes.

\noindent\textbf{Pose preprocessing}
Regarding the 2D pose, we utilize the dedicated 2D pose estimator for better accuracy, since direct estimation of 2D pose is more accurate compared to projection from estimated 3D pose.
The 2D pose is aligned based on the bounding box of the signer by applying the affine transform.
After then, we rescaled the 2D pose sequence from the bounding box into the [0, 1] range.
The 3D poses, including the positional pose, the rotational pose and the facial expression, we utilize the off-the-shelf 3D expressive human pose estimator Hand4Whole~\cite{moon2022accurate} as illustrated in Figure~\ref{fig:qualitative_result}.
For the 3D positional pose, we aligned the translation by placing the root joint of the first frame at the origin without normalization. 
In addition, the 3D rotational pose and the facial expression features are used without preprocessing.

\subsection{Ablation study}
We conduct an ablation study on two main contributions of our framework: part-wise motion context learning and pose ensemble.
For each experiment, the contribution that is not being tested remains identical to our proposed P3D.

\noindent\textbf{Effect of part-wise motion context learning.}
In Table~\ref{table:pose_encoding_methods}, we compare the results of various pose encoding methods in single iteration of P3D.
The MLP is the simple alternative to the proposed PET or WET, which is constructed of stacked fully-connected layers, batch normalization, and dropout.
Experimental results demonstrate that employing our PET and WET gives a large performance gain over MLP alternatives.

We also report that a combination of PET and WET is essential in sign language recognition.
When the number of classes is small (WLASL 100, WLASL 300), utilizing one of the WET (the fourth row) or PET (the third row) provides a similar result to utilizing both PET and WET (the fifth row).
However, when the number of classes is large (WLASL 2000), utilizing both PET and WET performs the best.
As the number of classes increases, learning highly distinguishable features from motion context becomes crucial.
For that, the PET and WET are dedicated to capturing intra-part context and merging them into a unified inter-part context, respectively.
Such design benefits the task-specific representation power on sign language recognition, that each part (\ie body, hands, and facial expressions) holds its own important motion context.
As a result, our PET-WET encoder outperforms PET-PET and WET-WET by 3.75\% and 12.08\%, respectively.

\noindent\textbf{Effect of each human part.}
In Table~\ref{table:input_human_parts}, we conduct an ablation study on employing different human part inputs.
As our task is sign language recognition, hands are the most essential part of accurate recognition.
However, the additional gain can be obtained by leveraging body pose and facial expression on top of hands.
It demonstrates that body pose and facial expression, as well as hands, also provide helpful information for sign language recognition.

\begin{figure}[t]
\begin{center}
\vspace{2mm}
\includegraphics[width=1.\linewidth]{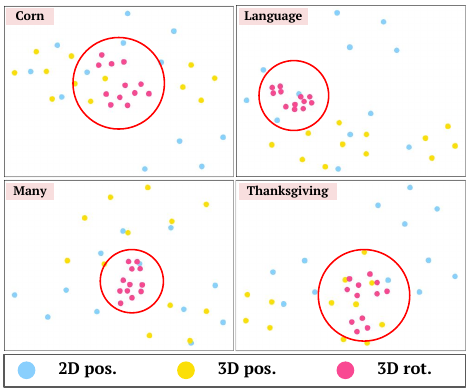}
\end{center}
\caption{\textbf{t-SNE~\cite{van2008visualizing} visualization of features from different pose representations.}
Each plot corresponds to a single word ("Corn", "Language", "Many", and "Thanksgiving", respectively), and each point is feature $X^{(N)}$ from each video.
All test videos of the corresponding word in WLASL 2000 are employed for the visualization.
}
\label{fig:tsne}
\end{figure}

\noindent\textbf{Effect of each pose representation.}
Table~\ref{table:pose_representations} shows that jointly utilizing 2D positional pose, 3D positional pose, and 3D rotational pose achieves the best performance. In terms of single representation, the 3D rotational pose is the most effective pose representation for sign language recognition.
Furthermore, additionally using the 3D rotational pose to other pose representations (the first row \& the fifth row, the second row \& the sixth row, and the fourth row \& last row) consistently improves the accuracy.
As a reason for the effectiveness of 3D rotational pose for sign language recognition, we argue that 3D rotational pose is less sensitive to signer variation than 2D positional pose and 3D positional pose.
Figure~\ref{fig:tsne} illustrates that features of single words from different signers are closely embedded when input pose representation is a 3D rotational pose.
On the other hand, in the case of the 2D and 3D positional poses, the embedded features of the single words show large discrepancies when the signers are different.
Such advantage of 3D rotational pose leads to improved performance over other pose representations since the sign languages in the WLASL dataset are signer-independent~\cite{li2020word}.

\noindent\textbf{Pose ensemble methods.}
In Table~\ref{table:pose_fusion}, we compare the performances of different pose ensemble methods.
Since we employ three types of pose representations (2D positional pose, 3D positional pose, and 3D rotational pose), how to ensemble information in each pose representation is an important issue.
We design three strategies for ensemble methods, and the description for each ensemble method is as follows.
\begin{itemize}
\item\textit{Early ensemble (Ours):}
All pose representations of each joint are concatenated before being fed to the PET-WET blocks.
Thus, PET-WET blocks the process motion context of all poses representations simultaneously.

\item\textit{Middle ensemble:}
Separate PET-WET blocks are utilized to extract feature $X^{(n)}$ for each pose representation: $X_\text{2D pos}^{(n)}$, $X_\text{3D pos}^{(n)}$, and $X_\text{3D rot}^{(n)}$.
The extracted features of all pose representations are averaged before being fed to the prediction head.

\item\textit{Late ensemble:}
Same as the middle ensemble, but each feature of each pose representation is fed to its own prediction head.
The output of all prediction heads is averaged to regress a single word.
\end{itemize}
The experimental results show that \textit{Early ensemble} performs best for the WLASL 2000.
In Table~\ref{table:pose_ensemble_flops}, we compare the computational costs of ensemble methods.
As separate PET-WET blocks are utilized in \textit{Middle ensemble} and \textit{Late ensemble}, \textit{Early ensemble} is most efficient in terms of the number of model parameters and FLOPs.
Considering the accuracy and efficiency, we adopt \textit{Early ensemble} as our ensemble strategy.

\subsection{Comparison with state-of-the-art methods}
Table~\ref{table:sota_comparision} shows that P3D outperforms previous pose-based methods.
We first demonstrate the capability of our proposed part-wise motion context learning model via quantitative comparison on the WLASL dataset using only 2D pose.
Then we show that 3D pose fusion is more effective in our P3D than in other models, leading to our model outperforming others on the WLASL dataset using 2D and 3D pose.

\noindent\textbf{Using only 2D pose.}
The second block of Table~\ref{table:sota_comparision} shows that our P3D outperforms existing methods on WLASL 300 and WLASL 2000 split with 2D pose input.
In this experiment, we only use 2D pose for the fair comparison of models to learn highly distinguishable features from motion context.
Prior to our work, state-of-the-art methods were SPOTER~\cite{bohavcek2022spoter} in WLASL 100, and ST-GCN~\cite{yan2018stgcn} in WLASL 300 and WLASL 2000.
From the benefits of its simple and intuitive Transformer architecture, SPOTER achieved the best performance on WLASL 100.
However, due to the limitation on representation power, SPOTER did not perform well on more complex problems such as WLASL 300 and WLASL 2000.
Meanwhile, graph convolution via spatio-temporal adjacency used in ST-GCN is observed to be effective on a large number of classes (WLASL 300 and WLASL 2000).
Yet, employing the spatio-temporal adjacency graph leads to two cons: it cannot model the long-term dependency, and cannot learn the part-specific context.
In contrast, the proposed part-aware Transformer architecture of our P3D is capable of handling the long-term dependency and part-specific context at the same time.
As a result, our P3D shows better performance compared to ST-GCN on every split of WLASL, proving the better representation power of proposed part-wise motion context learning.

\noindent\textbf{Jointly using 2D and 3D pose.}
The third block of Table~\ref{table:sota_comparision} shows that our P3D scores highly superior results compared to previous methods with using both 2D and 3D poses as an input.
Again, in order to conduct fair comparison, identical input features are used and the same pose ensemble method is used for every tested models.
We performed a comparative analysis in two aspects as follows.

First, our P3D resulted in a much higher performance gain from jointly using 2D and 3D poses.
While our P3D gained 6.39\% top-1 per-instance recognition accuracy on WLASL2000 by pose ensemble, gains of previous state-of-the-art methods SPOTER~\cite{bohavcek2022spoter} and ST-GCN~\cite{yan2018stgcn} are confined to 2.05\% and 2.97\%, respectively.
Such observation suggests that our part-wise motion context learning is more capable of acquiring contextual information from various types of pose representations, which distinguishes it from the basic Transformer architecture of SPOTER~\cite{bohavcek2022spoter} or the spatio-temporal graph convolution of ST-GCN~\cite{yan2018stgcn}.

Second, P3D achieved superior performance compared to existing methods, particularly by a large margin on a large number of classes such as WLASL300 and WLASL2000.
As demonstrated in the third block of Table~\ref{table:sota_comparision}, our P3D outperformed previous pose-based methods SPOTER~\cite{bohavcek2022spoter} and ST-GCN~\cite{yan2018stgcn} on every subset of WLASL.

\noindent\textbf{Using 3D expressive pose.}
The fourth block of Table~\ref{table:sota_comparision} shows the performance with full input features: 2D pose, 3D positional and rotational pose, and facial expression.
With additional usage of facial expression over the third block of Table~\ref{table:sota_comparision}, we observed the performance gain of 0.79\%, 0.93\%, and 2.12\%, respectively for each subset of WLASL.
Moreover, when compared to the RGB-based models presented in the first block of Table~\ref{table:sota_comparision}, our P3D scored comparable results in WLASL100 and WLASL300, and outperformed in WLASL2000.
Our framework is the first to outperform (WLASL 2000) or achieve comparable performance (WLASL 100, WLASL 300) with RGB-based methods.

\section{Conclusion}
In this work, we proposed the sign language recognition framework P3D by leveraging part-wise 3D motion context learning.
We utilize the alternating PET and WET layers, specializing in learning intra-part and inter-part motion contexts, respectively.
Learning part-wise motion context resulted in superior performance on the WLASL benchmark compared to previous methods.
Moreover, we jointly utilized 2D and 3D poses via joint-wise pose concatenation to fully exploit the performance of pose-based sign language recognition.
We observed that joint-wise concatenation has advantages on both sides of performance and efficiency.
As a result, our P3D outperformed the previous methods by a large margin on WLASL, jointly using 2D and 3D poses.

\section*{Acknowledgements}
This work was supported in part by the IITP grants [No.2021-0-01343, Artificial Intelligence Graduate School Program (Seoul National University), No. 2021-0-02068, and No.2023-0-00156], the NRF grant [No. 2021M3A9E4080782] funded by the Korea government (MSIT), and the SNU-Samsung SDS Research Center.

{\small
\bibliographystyle{ieee_fullname}
\bibliography{main}
}

\end{document}